\documentclass[10pt,twocolumn,letterpaper]{article}

\usepackage[pagenumbers]{cvpr} 

\usepackage{graphicx}
\usepackage{amsmath}
\usepackage{amssymb}
\usepackage{booktabs}

\usepackage[pagebackref,breaklinks,colorlinks]{hyperref}

\usepackage[accsupp]{axessibility}  

\usepackage[capitalize]{cleveref}
\crefname{section}{Sec.}{Secs.}
\Crefname{section}{Section}{Sections}
\Crefname{table}{Table}{Tables}
\crefname{table}{Tab.}{Tabs.}

\def\cvprPaperID{6666} 
\def\confName{CVPR}
\def\confYear{2024}

\begin{document}

\title{Exploring Facial Expression Recognition through Semi-Supervised Pretraining and Temporal Modeling}

\author{
Jun Yu$^1$, Zhihong Wei$^1$, Zhongpeng Cai $^1$\thanks{Corresponding author}, Gongpeng Zhao$^1$, Zerui Zhang$^1$, Yongqi Wang$^1$,\\ Guochen Xie$^1$, Jichao Zhu$^1$, Wangyuan Zhu$^1$  \\
$^1$University of Science and Technology of China\\
\tt\small harryjun@ustc.edu.cn\\
\tt\small \{weizh588,zpcai,zgp0531,igodrr,wangyongqi,xiegc,\\
\tt\small jichaozhu,zhuwangyuan\}@mail.ustc.edu.cn 
}


\maketitle

\begin{abstract}
Facial Expression Recognition (FER) plays a crucial role in computer vision and finds extensive applications across various fields. This paper aims to present our approach for the upcoming 6th Affective Behavior Analysis in-the-Wild (ABAW) competition, scheduled to be held at CVPR2024. In the facial expression recognition task, The limited size of the FER dataset poses a challenge to the expression recognition model's generalization ability, resulting in subpar recognition performance. 
To address this problem, we employ a semi-supervised learning technique to generate expression category pseudo-labels for unlabeled face data. At the same time, we uniformly sampled the labeled facial expression samples and implemented a debiased feedback learning strategy to address the problem of category imbalance in the dataset and the possible data bias in semi-supervised learning. Moreover, to further compensate for the limitation and bias of features obtained only from static images, we introduced a Temporal Encoder to learn and capture temporal relationships between neighbouring expression image features. In the 6th ABAW competition, our method achieved outstanding results on the official validation set, a result that fully confirms the effectiveness and competitiveness of our proposed method.
\end{abstract}

\section{Introduction}
\label{sec:intro}
The goal of Facial Expression Recognition (FER) is to identify the emotional state of an individual by analyzing facial images or videos. It is a broad research field spanning multiple fields such as machine learning, image processing, psychology, etc., with a wide range of applications, including safe driving, intelligent monitoring, and human-computer interaction. In view of its diverse applications, it is extremely important to establish a robust FER system. The expression recognition task is a classic problem in the field of pattern recognition, which usually involves the classification of six basic emotions: happiness, surprise, sadness, anger, disgust and fear.
The facial expression recognition community has made tremendous progress in recent years. State-of-the-art FER methods achieve very good results on numerous public datasets, such as RAF-DB\cite{li2017reliable}, SFEW\cite{yu2015image} and AffectNet\cite{mollahosseini2017affectnet}.

To stimulate interdisciplinary collaboration and address pivotal research inquiries spanning affective computing, machine learning, and multi-modal signal processing, Kollias et al. have spearheaded the Affective Behavior Analysis in-the-wild (ABAW) initiative. The 6th ABAW workshop and competition is slated to align with the IEEE CVPR conference in 2024\cite{zafeiriou2017aff,kollias2019deep, kollias2019face, kollias2019expression, kollias2020analysing,kollias2021affect,kollias2021analysing,kollias2021distribution,kollias2022abaw, kollias2023abaw,kollias2023abaw2,kollias20246th}.

Traditional fully supervised Facial Expression Recognition (FER) methods rely on the availability of large volumes of high-quality labeled data to fine-tune model precision \cite{tian2011facial, chen2021understanding, chang2021learning}. However, mainstream training datasets often suffer from class imbalance. Most fully supervised models tend to accurately identify majority classes, thus reducing the model's accuracy for minority classes. Sometimes minority classes constitute less than 10\% of the data, and this disparity in data volume makes it challenging for models to learn fairly across all categories. Financial and logistical constraints in acquiring extensive labeled FER data hinder the expansion of training repositories. In contrast, the volume of data for Face Recognition (FR) surpasses that of FER. Making a leap to augment samples from FR data to aid models in learning the FER task could be highly beneficial. Addressing the class imbalance issue and the discrepancy in data distribution between FR and FER data to effectively utilize FR data in support of FER models presents an urgent challenge. 

In this study, we propose a two-phase methodology to enhance the recognition and analysis of facial expressions. The first phase, known as the spatial pre-training phase, plays a crucial role in preparing the model for subsequent tasks. This phase is an improvement on the method\cite{zeng2022face2exp}. During this phase, we leverage the power of semi-supervised learning techniques to generate pseudo-labels for expression categories using unlabeled face data. This approach ensures a sufficiently large training corpus, allowing the model to effectively extract robust facial expression features. To tackle the challenges of category imbalance in the dataset and potential data bias during semi-supervised learning, we adopt two strategies. First, we uniformly sample labeled facial expression instances to address the category imbalance. Additionally, we employ a debiased feedback learning strategy to mitigate the impact of potential data bias. These strategies collectively contribute to training a more robust facial expression recognizer. Moving to the second phase, the Temporal Refine phase, we aim to further improve the recognition and analysis of facial expressions by capturing the temporal dynamics. In this phase, we freeze the facial expression recognizer trained in the first phase, which has already acquired strong spatial representation capabilities. To incorporate temporal information, we introduce a temporal encoder that learns the temporal relationships between neighbouring expression image features. By considering the temporal aspect, we compensate for the inherent feature bias obtained solely from static images. This integration of temporal dynamics enables more accurate and comprehensive dynamic recognition and analysis of facial expressions.

To sum up, our contributions  can be summarized as:
\begin{itemize}
    \item To address the problem of scarcity of facial expression data, we applied a semi-supervised learning technique to generate expression category pseudo-labels for unlabeled face data. At the same time, we uniformly sampled the labeled facial expression samples and implemented a debiased feedback learning strategy to solve the problem of category imbalance in the dataset and the possible data bias in semi-supervised learning.
    \item In order to compensate for the limitations and biases of features acquired only from static images, we introduce a temporal encoder to learn and capture temporal relationships between neighbouring expression image features. This strategy aims to enhance the model's ability to recognize and analyze the dynamic changes in facial expressions and achieve more accurate dynamic facial expression recognition.
    \item In the 6th ABAW competition, our algorithm achieved outstanding results on the official validation set, a result that fully confirms the effectiveness and competitiveness of our proposed method.
\end{itemize}

\begin{figure*}[ht]
\centering
\includegraphics[width=0.8\textwidth]{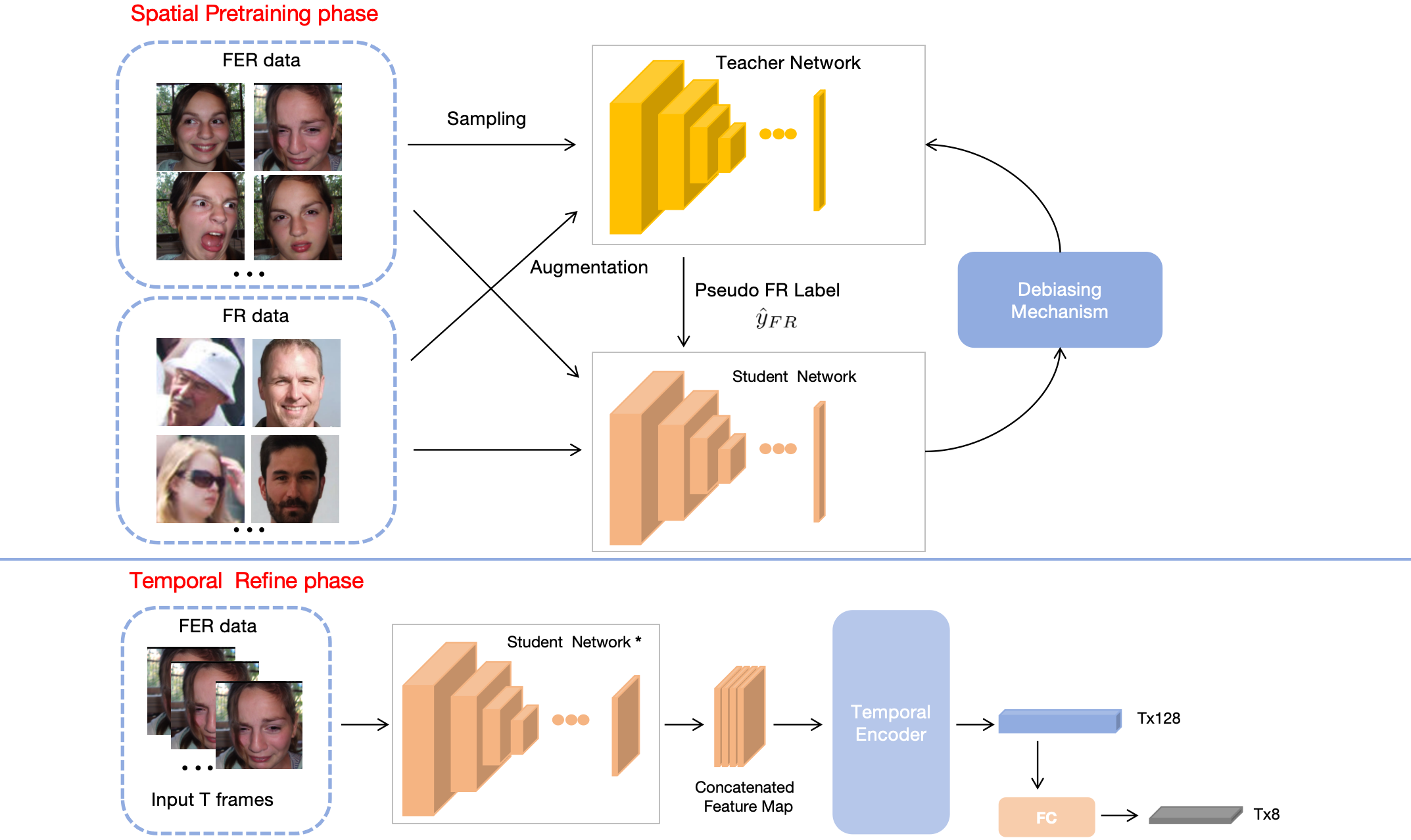}
\caption{Framework Description. Our approach is mainly divided into a Spatial Pretraining phase and a Temporal Refine phase. (1). The goal of the Spatial Pretraining phase is to expand face expression data by mining large-scale unlabeled faces through a semi-supervised algorithm. (2). The goal of the Temporal Refine phase is to do temporal feature enhancement of the image features extracted by the student network in the first phase by means of a temporal encoder, so as to improve the accuracy of recognizing the dynamic facial expressions in the video.
}
\label{fig:Pipeline}
\end{figure*}

\section{Related Work}
\label{sec:Relatedwork}

\subsection{Facial Expression Recognition}

The task of recognizing facial expressions is a foundational challenge in pattern recognition. Techniques that leverage fully supervised data have significantly advanced the field of Facial Expression Recognition (FER), as evidenced by groundbreaking research\cite{wen2021distract, she2021dive,  zhang2022learn,chen2021understanding,xue2021transfer}. However, datasets for facial expressions face notable limitations, including a lack of diversity and significant category imbalances. In response, recent efforts have shifted towards expanding these datasets to enrich the variety of facial expressions available for analysis. One pioneering approach to address the challenge of inconsistent labeling across different facial expression datasets is the IPA2LT framework \cite{zeng2018facial}. This methodology introduced the LTNet scheme, an innovative strategy for uncovering the underlying truths among diverse, inconsistent labels through the use of embedded analysis.In the realm of semi-supervised learning for FER, Ada-CM\cite{li2022towards} emerged as the first to investigate the concept of dynamic confidence. By designing an adaptive confidence margin, this approach innovatively adapts during training to maximize learning from unlabeled data through feature-level comparisons, utilizing the InfoNCE loss\cite{wu2021rethinking} to capture valuable features effectively.Further advancing the field, Face2Exp\cite{zeng2022face2exp} introduced the Meta-Face2Exp framework. This methodology utilizes a meta-optimization framework to derive unbiased knowledge from auxiliary Facial Recognition (FR) data, showcasing a novel approach to knowledge extraction.
Lastly, an innovative technique proposed by Zhang et al.\cite{zhang2024leave} employs rebalancing attention mapping to regularize models. This allows for the extraction of transformation-invariant information from secondary categories across all training samples, addressing the critical issue of data imbalance by focusing on underrepresented categories.

\subsection{Semi-Supervised Learning}

Semi-supervised learning, focusing on unlabeled data, is a vital area within learning algorithms. A prominent approach in Semi-Supervised Learning is generating synthetic labels for unlabeled images and guiding the algorithm to identify these labels for such image inputs, as discussed in the works \cite{lee2013pseudo,xie2020self}. Similarly, consistent regularization, according to \cite{bachman2014learning,laine2016temporal,sajjadi2016regularization} utilizes the model’s predictive capacity to form a synthetic label by altering the input or model’s operations randomly. Fixmatch\cite{sohn2020fixmatch} integrates the benefits of generating synthetic labels and applying both light and intensive data augmentations to ensure consistent regularization and generate pseudo-labeled data from samples exceeding a confidence threshold. However, its limitation arises early in model training due to the static threshold.To mitigate this, Flexmatch\cite{zhang2021flexmatch} proposes Curriculum Pseudo Labeling (CPL), leveraging unlabeled data based on the model's learning stage and dynamically adjusting thresholds for various classes, thus allowing the inclusion of informative unlabeled data and their pseudo labels. Additionally, Dash \cite{xu2021dash} employs a dynamic threshold to keep only the samples with losses under a specific threshold during updates.

\section{Method}

\subsection{Overview}
As shown in Figure.\ref{fig:Pipeline}, our approach uses multiple techniques to improve facial expression recognition (FER) in two phases. The first phase, the Spatial Pretraining phase, uses semi-supervised learning to increase the FER dataset by using facial recognition (FR) data. This helps in training a better image feature extractor. We use two neural networks, a teacher and a student, with the same structure but different weights. In this phase, we improve the networks by using a debiasing method. This involves comparing biased FR data (unlabeled) with debiased FER data (labeled) to create pseudo-labels. The teacher network learns from balanced FER data and makes these pseudo-labels, which the student network then uses to learn from the FR data, adjusting based on feedback to improve accuracy. In the training, the teacher and student networks update alternately with the debiasing method to better process FR data, improving accuracy and debiasing.
The second phase, the Temporal Refine phase, uses the student network (fixed after the first phase) to extract image features. To address biases from static images, we use a temporal encoder to understand time-related feature relationships, improving system accuracy. This enables dynamic recognition and analysis of facial expressions in videos, with a classifier providing frame-by-frame predictions.

\subsection{Data Pre-process}
\subsubsection{Data Sampling}
To achieve class balance, we meticulously sample the labeled Facial Expression Recognition (FER) data, ensuring an equal distribution of samples across each expression category. This strategic sampling method allows the model to learn features that are more evenly distributed among classes, significantly contributing to the de-biasing process in Facial Recognition (FR) data analysis.

\subsubsection{Data Augmentation}
To enhance the diversity and robustness of our dataset, we implement two distinct data augmentation strategies. In the case of Weakly-Augmented (WA) operations, our approach primarily encompasses horizontal flipping, color jittering, and similar techniques to introduce subtle variations. On the other hand, for Strongly-Augmented (SA) operations, we adopt RandAugment\cite{cubuk2020randaugment}, a method known for its efficacy in generating more significant transformations while maintaining the integrity of the data. These augmentation strategies are meticulously chosen to ensure that, despite the alterations, the unlabeled samples retain their semantic consistency.

\subsection{Semi-Supervised Training}

The focus of this phase is to utilize semi-supervised learning to extend the FER dataset by using the Facial Recognition (FR) dataset. This expansion helps train an efficient static image feature extractor. In this phase, we employ two neural networks with identical structures but independent weights: the teacher network (T) and the student network (S).

To optimize network performance in this phase, we employ a debiasing mechanism as a core strategy. This involves analyzing the disparities between biased FR data (which is unlabeled) and debiased FER data (which is labeled) to generate pseudo-labels. By sampling FER data in a category-balanced manner, the teacher network learns and generates these pseudo-labels. Subsequently, the student network employs these pseudo-labels to train on the FR data, continuously adjusting itself based on feedback to enhance recognition accuracy. During the training process, the teacher network and the student network are alternately updated, coupled with the de-biasing strategy. This iterative updating aims to improve both the accuracy and de-biasing effect of processing FR data. Consequently, this enhances the prediction ability of the student network.

\subsubsection{Student Network}
To enhance FER, the student network leverages the rich and comprehensive diversity of large-scale unlabeled FR data. During training, the student network utilizes this unlabeled FR data along with pseudo-labels generated by the teacher's network. The training process involves encouraging both networks to predict similar conditional classification distributions for the unlabeled FR data, achieved through the use of the $\mathcal{L}_u$ loss function. The expression for $\mathcal{L}_u$ is given as:

\begin{equation}
    \mathcal{L}_u=\mathrm{CE}(\hat{y}_{FR},\mathcal{S}(x_{FR};\theta_{\mathrm{s}})).
\end{equation}

where $\theta_{\mathrm{s}}$ is the parameter of the student network, and the $\mathrm{CE}$ denotes the cross-entropy loss.

As the student network undergoes updates, the pseudo-labeling dynamically changes throughout the training process. The parameters $\theta_{\mathrm{s}}$ of the student network are updated during the meta-training phase. In the meta-testing phase, the balanced FER dataset (which was used to train the teacher network) is employed to estimate the cognitive differences between the biased FR data and the de-biased FER.

\subsubsection{Teacher Network}
A sampling module, denoted as $\mathrm{Smp(\cdot)}$, is utilized to guarantee a balanced class distribution in the Facial Expression Recognition (FER) dataset. To achieve this, an equal number of samples from each facial expression category were randomly selected, thereby ensuring a balanced class representation for training the teacher network. The learning process integrates three specific types of loss functions: supervised loss, consistency loss, and feedback loss, which collectively provide effective guidance to the teacher network. This process is represented by the following equation:

\begin{equation}
\mathcal{L}_{\mathcal{T}}=\mathcal{L}_s+\mathcal{L}_c+\mathcal{L}_f.
\end{equation}

Specifically, the supervision loss and consistency loss apply only to the teacher network, while the feedback loss takes into account the predictive distribution of the meta-test for the student network.

For the  supervised learning, aimed at minimizing the supervised loss function on a balanced and labeled Facial Expression Recognition (FER) dataset, the supervised loss function $\mathcal{L}_s$ is formulated as follows:
\begin{equation}
\mathcal{L}_s=\mathsf{C}\text{E}(y_{FER},\mathcal{T}(x_{FER};\theta_{\mathrm{t}})).
\end{equation}
where $\theta_{\mathrm{t}}$ is a parameter of the teacher network and $\mathsf{C}\text{E}$ denotes the cross-entropy loss.

For consistency learning,the teacher network requires that the original image and the augmented counterpart have close class-conditional distributions with a consistency loss function $\mathcal{L}_c$:
\begin{equation}
\mathcal{L}_c=\mathrm{CE}(\mathcal{T}(x_{FR};\theta_{\mathrm{t}}),\mathcal{T}(\mathrm{Aug}(x_{FR});\theta_{\mathrm{t}})).
\end{equation}

The $\mathrm{Aug(\cdot)}$ denotes strong data augmentation, this approach utilizes intensive data augmentation techniques, including rotation, removal, and pixel-level image manipulation.

For the feedback learning, the process involves estimating feedback based on cognitive discrepancies between the FR (Facial Recognition) and FER (Facial Expression Recognition) datasets. This feedback is utilized to refine the parameters of the teacher network, with the feedback loss function $\mathcal{L}_f$ denoted by: 
\begin{equation}
\mathcal{L}_f=f\cdot\mathrm{CE}(\hat{y}_{FR},\mathcal{T}(x_{FR};\theta_{\mathrm{t}})).
\end{equation}
where the definition of the feedback coefficient $f$ can be expressed as:
\begin{equation}
\begin{aligned}f=\eta_{\mathcal{S}}\cdot(\nabla_{\theta_{\mathcal{S}}^{(t+1)}}\mathrm{CE}(y_{FER},\mathcal{S}(x_{FER};\theta_{\mathcal{S}}^{(t+1)}))^{\top}\cdot\\\nabla_{\theta_{\mathcal{S}}}\mathrm{CE}(\hat{y}_{FR},\mathcal{S}(x_{FR};\theta_{\mathcal{S}}^{(t)}))).
\end{aligned}
\end{equation}
The $f$ is denoted as the dot product of two terms. First term: the gradient of the new student network on the debiased FER data. Second term: gradient of the old student network on biased FR data

\subsubsection{Debiasing Mechanism}

The performance of a novel student network on balanced Facial Expression Recognition (FER) data serves as the criterion for evaluation. More precisely, positive feedback coefficients are achieved (indicating a favourable $\mathcal{L}_f$ value) when both the student network operating on Face Recognition (FR) data and the novel student network focused on FER data share identical gradient orientations. This scenario promotes the modification of the teacher network through the employment of the gradient's current trajectory. Conversely, should the student network concerning FR data and the novel student network for FER data exhibit contrasting gradient orientations, the feedback coefficients assume a negative value. This acts as a deterrent to the teacher network's adjustment, advocating for the use of an opposing gradient direction. Thus, feedback operates as a crucial reward signal, retroactively influencing the teacher network by dictating the influence of its parameters on the student network's gradient for the extraction of unbiased features.

\subsection{Temporal encoder}

The Transformer-based Temporal Encoder is a novel neural network architecture that we have specifically developed to capture temporal relationships between image features, aiming to address feature biases obtained solely from static images. This design leverages the self-attention mechanism of the Transformer architecture to effectively process sequence data and learn temporal dependencies within image sequences. By applying self-attention in the temporal dimension, the Temporal Encoder enables the model to emphasize keyframes and features in an image sequence, facilitating the extraction of spatio-temporal features.

\subsection{Post-Process}
Given that the Aff-Wild2 dataset is constructed from consecutive frames of videos, and considering that the formation of facial expressions unfolds over a period of time, it follows logically that significant alterations in facial expressions are unlikely to occur within a narrow sequence of adjacent frames. In light of this, we implemented a sliding window technique to refine the predictive outcomes, aiming to enhance the consistency of the expression labels. This process involves aggregating the frequencies of all predicted labels within each window. Subsequently, the label that predominates in frequency within a given window is designated as the representative expression for all frames encompassed by that window. By applying this sliding window approach across the entire dataset, we effectively facilitate the smoothing of predicted expression labels, thereby achieving a more coherent and accurate representation of facial expressions throughout the dataset.

\begin{table*}
  \centering
  \caption{Ablation study results on the validation set.}
  \label{tab:table1}
  \resizebox{\linewidth}{!}{
  \begin{tabular}{ccccccccccccccc}
    \toprule
    \textbf{Method} & \textbf{Aff-Wild2} & \textbf{AffectNet and ExpW} & \textbf{MS1MV2} &\textbf{ Temporal Encoder }& \textbf{Post-Process} & \textbf{F1 Score (\%)} \\ 
    \midrule
	Baseline & $\checkmark$ & &  &  &  &23.00 \\
	SSL & $\checkmark$ &  &$ \checkmark$ & &  &39.96  \\

        SSL & $\checkmark$ & $\checkmark$ & \checkmark & &  &40.57  \\

	SSL + Temporal & $\checkmark$ & $\checkmark$  & $\checkmark$  & $\checkmark$ & &42.77 \\
 
	SSL + Temporal+ Post-process & $\checkmark$ & $\checkmark$  & $\checkmark$ & $\checkmark$ &  $\checkmark$  &44.43  \\

    \bottomrule
  \end{tabular}
  }
\end{table*}

\section{Experiment}
In this section, we will provide a detailed description of the used datasets, the experiment setup, and the experimental results.
\subsection{Datasets}
\label{sec:dataset}
\textbf{FER Datasets.}

The 6th Workshop and Competition on Affective Behavior Analysis in-the-wild unveiled the Aff-wild2 database, a pivotal collection drawn from a range of studies\cite{zafeiriou2017aff,kollias2019deep, kollias2019face, kollias2019expression, kollias2020analysing,kollias2021affect,kollias2021analysing,kollias2021distribution,kollias2022abaw, kollias2023abaw,kollias2023abaw2,kollias20246th}. This resource was central to the EXPR Classification Challenge, featuring an audio-visual dataset comprised of 548 videos and approximately 2.7 million frames annotated for six fundamental facial expressions, as well as neutral and 'other' categories. To enrich the dataset's depth, we incorporated data from the AffectNet and ExpW databases; AffectNet contributed around 1 million facial images categorized into 11 emotions, while ExpW offered 91,793 images across seven expression categories. This integration aimed to enhance the comprehensiveness and utility of our analysis.

To further enhance the diversity of our dataset and bolster the generalizability of our model, we also randomly selected 8,000 images for each category from a merged dataset comprising AffecNet and ExpW. For images falling into the 'other' category, we relied exclusively on the Aff-Wild2 dataset, given the absence of such images in the alternate databases. The compilation is rounded off with the inclusion of the remaining images from these datasets as unlabeled samples, serving to augment our dataset's comprehensiveness.

\textbf{FR Datasets.}

In our work on Face Recognition Datasets, we employ the MS1MV2 dataset\cite{huang2008labeled} as the source of unlabeled data. This dataset, a semi-automatically refined iteration of the MS-Celeb-1M dataset\cite{guo2016ms}, was developed by ArcFace\cite{deng2019arcface} and contains approximately 85,000 identities and 5.8 million images. For the purposes of our experiments, we utilized a subset of the MS1MV2 dataset, curated from InsightFace\cite{insightface}, by uniformly selecting one-third of the images. This resulted in a comprehensive subset encompassing 1.94 million images.

\subsection{Setup}

All training face images are detected and resized to 256$\times$256 pixels, and augmented by random cropping to 224$\times$224 pixels. ResNet50 is selected by default to serve as the core framework for both the teacher and the student networks. Initially, the learning rates are determined to be 1 $\times$ 10$^{-2}$ for the teacher network and 1 $\times$ 10$^{-3}$ for the student network, with subsequent adjustments made through a technique known as cosine annealing. A batch size of 32 is established. This training regimen unfolds over 100,000 steps and It is trained end-to-end with one Nvidia V100 GPU. For timing fine-tuning, the input video is 64 frames as a clip, and a window size of 30 is used for post-processing.

\subsection{Metrics}
In line with the competition requirements, we employ the average F1 score as our evaluation metric, which is robust against class frequency variations and particularly suitable for imbalanced class distributions. The calculation of the average F1 score is as follows:

\begin{equation}
  F_1^c = \frac{2\times Precision\times Recall}{Precision + Recall}
\end{equation}
\begin{equation}
   F1 = \frac{1}{N}\sum_{c=1}^{N} F_1^c
\end{equation}
where \begin{math}N \end{math} represents the number of classes and \begin{math}c \end{math} means $c$-th class.

\subsection{Results}

\label{sec:ablation study}

To validate the efficacy of our approach, we executed ablation studies on each component and strategy within our method, with the outcomes presented in Table.\ref{tab:table1} It is evident that the application of the SSL (Semi-Supervised Learning) technique significantly enhances the recognition performance by 16.96\%. Further augmentation of the facial expression dataset can lead to an increase in accuracy. Moreover, incorporating the temporal encoder results in an additional 2.2\% improvement in accuracy, underscoring the importance of temporal learning. Ultimately, through post-processing, the model achieves an accuracy rate of 44.43\%.

\section{Conclusion}
In this paper, we propose a two-phase approach to improve facial expression recognition. The first phase is called the spatial pre-training phase, in which, in order to address the problem of scarcity of facial expression data, we employ a semi-supervised learning technique to generate expression category pseudo-labels for unlabeled facial data. At the same time, we uniformly sampled the labeled facial expression samples and implemented a debiased feedback learning strategy to address the problem of category imbalance in the dataset and possible data bias in semi-supervised learning. The second phase is the temporal refine phase. In this phase, to compensate for the limitation and bias of obtaining features only from static images, we introduced a temporal encoder to learn and capture the temporal relationship between the features of neighbouring expression images to achieve more accurate dynamic facial expression recognition. 

{\small
\bibliographystyle{ieee_fullname}
\bibliography{egbib}
}

\end{document}